\documentclass[sigconf]{acmart}
\AtBeginDocument{%
  }

\usepackage{graphicx}
\usepackage{amsmath}
\usepackage{amsthm}
\usepackage{booktabs}
\usepackage[switch]{lineno}
\usepackage{amssymb}
\usepackage{multirow}
\usepackage[linesnumbered,ruled,vlined]{algorithm2e}
\usepackage{amsmath} 
\usepackage{xcolor}
\usepackage{setspace}
\usepackage{wrapfig} 
\newtheorem{theorem}{Theorem}[section]
\newtheorem{assumption}{Assumption}[section]
\newtheorem{lemma}{Lemma}[section]
\newtheorem{definition}{Definition}[section]
\newtheorem{proposition}{Proposition}[section]

\setcopyright{acmlicensed}
\copyrightyear{2025}
\acmYear{2025}
\setcopyright{acmlicensed}\acmConference[ICAIF '25]{6th ACM International Conference on AI in Finance}{November 15--18, 2025}{Singapore, Singapore}
\acmBooktitle{6th ACM International Conference on AI in Finance (ICAIF '25), November 15--18, 2025, Singapore, Singapore}
\acmDOI{10.1145/3768292.3770349}
\acmISBN{979-8-4007-2220-2/2025/11}

\acmSubmissionID{66}

\begin{document}

\title[LENS]{LENS: Large Pre-trained Transformer for Exploring Financial Time Series Regularities}

\author{Yuanjian Xu}
\email{yxu085@connect.hkust-gz.edu.cn}
\affiliation{%
  \institution{The Hong Kong University of Science and Technology (Guangzhou)}
  \city{Guangzhou}
  \country{China}
}
\author{Jianing Hao}
\email{jhao768@connect.hkust-gz.edu.cn}
\affiliation{%
  \institution{The Hong Kong University of Science and Technology (Guangzhou)}
  \city{Guangzhou}
  \country{China}
}
\author{Anxian Liu}
\email{aliu789@connect.hkust-gz.edu.cn}
\affiliation{%
  \institution{The Hong Kong University of Science and Technology (Guangzhou)}
  \city{Guangzhou}
  \country{China}
}
\author{Zhenzhuo Li}
\email{zli743@connect.hkust-gz.edu.cn}
\affiliation{%
  \institution{The Hong Kong University of Science and Technology (Guangzhou)}
  \city{Guangzhou}
  \country{China}
}
\author{Shichang Meng}
\email{shichmeng2-c@my.cityu.edu.hk}
\affiliation{%
  \institution{City University of Hong Kong}
  \city{Hong Kong}
  \country{China}
}

\author{Shuai Yuan}
\email{yuanshuai@stu.pku.edu.cn}
\affiliation{%
  \institution{Peking University}
  \city{Beijing}
  \country{China}
}
\author{Guang Zhang}
\email{guangzhang@hkust-gz.edu.cn}
\affiliation{%
  \institution{The Hong Kong University of Science and Technology (Guangzhou)}
  \city{Guangzhou}
  \country{China}
}

\renewcommand{\shortauthors}{Xu et al.}

\begin{abstract}
Modeling large-scale time series has gained significant attention in recent years. 
However, its direct application in finance remains challenging due to substantial differences in data characteristics across domains. 
Specifically, financial systems feature inherent stochasticity and low signal-to-noise ratios, rendering traditional methods and pre-training approaches ineffective.
This underscores the urgent need for a foundation model tailored to financial time series. 
To bridge this gap, we propose \textbf{LENS}, a pre-trained model for this domain.
\textbf{LENS} effectively captures the complexity of financial stochastic systems through a carefully crafted model architecture and mitigates noise during pre-training by using an invertible embedding module.
We provide a rigorous theoretical explanation of the model's effectiveness and validate its performance through extensive experiments. Pre-trained on a dataset comprising 100 billion financial observations, \textbf{LENS} achieves exceptional results across a wide range of critical downstream tasks.
Moreover, our work offers practical insights into developing pre-trained time series models in high-noise environments, paving the way for further advancements in this pivotal research domain. 
\end{abstract}




\maketitle

\section{Introduction}
Financial time series analysis remains a research focus for its critical role in decision-making processes, influencing strategies, risk assessments, and policy formulations~\cite{finance_survey_2020}. 
Unlike time series with clear trends and periodic patterns, like traffic flow data~\cite{trafficflow_2023}, financial time series are inherently random with low signal-to-noise ratios~\cite{character_2011,finance_ml_survey_2022}, driven by complex factors like economic conditions, political events, and market speculators~\cite{intro_to_ff_1996,trading_1994}, making them highly volatile and challenging to model effectively~\cite{stockpattern_2017}.

Previous research on modeling financial time series has explored a range of approaches, from traditional statistical methods to deep learning-based models~\cite{dependence_2006,ARFIMA-LSTM_2020}. 
While these methods often achieve satisfactory in-sample performance, their generalization ability remains limited, constrained by the expressive capacity of the models~\cite{stockpattern_2017}. 
The success of large language models (LLMs) in handling sequential data offers a promising new avenue for time series modeling~\cite{GPT4_2023,templlm_2023}. With their vast parameter capacity, LLMs significantly enhance modeling capabilities~\cite{fingpt_2023}. In recent years, efforts have been made to develop general-purpose large time series models capable of achieving strong performance across diverse domains~\cite{patchtst,tempo_2023}. However, unlike textual data, time series data vary significantly across domains, and data from other fields provide limited direct benefits for financial time series modeling~\cite{character_2011,fingpt_2023}.

Past pretraining experiences are challenging to directly transfer to financial time series foundation models due to the low signal-to-noise ratio and high randomness inherent in financial time series data. 
Thus, there is an urgent need for a dedicated financial time series foundation model and a scientifically designed training framework.
Our contributions are summarized as follows: We introduce \textbf{LENS}, a foundational model trained on over 100 billion financial data points. 
By leveraging diverse financial datasets, \textbf{LENS} demonstrates superior generalization across various downstream tasks, establishing itself as a robust foundation for financial time series modeling. 
Additionally, our architecture incorporates an invertible embedding module to initialize time series token representations at the patch level and a specialized attention mechanism for multivariate time series. 
These components effectively address the challenges of low signal-to-noise ratio and high randomness, capturing meaningful dependencies across variables.
Through comprehensive theoretical and experimental analysis, we validate the superior generalization performance of our architecture as a foundational building block for financial time series. 
\vspace{-0.3em}
\section{Methods}

In this section, we first detail the architecture of \textbf{LENS}.
As shown in Figure~\ref{fig:arch}, the model begins with a patching-based, invertible embedding module~\cite{patchtst}, which transforms input sequences into structured latent representations while preserving temporal information. 
\textbf{LENS}'s core is an encoder-decoder architecture composed of several TimeFormer blocks, augmented with time-aware and channel-aware attention mechanisms to better capture temporal dependencies and cross-channel interactions. We then describe the training procedure used to optimize the model parameters.

\subsection{Invertible Embedding Module}
A robust embedding module is essential for effective pretraining, particularly in noisy environments. 
Our design follows two core objectives: (i) ensuring the representations of input patches remain stable and discriminative under stochastic perturbations; and (ii) preserving invertibility for accurate original input reconstruction. 
To formalize robustness to noise, we make the following assumptions on the data generation process and the embedding function:
\begin{assumption}[Noisy Input Model]
The input samples $x_i$ and $x_i^+$ consist of clean signals $x_i^*$ and $x_i^{+*}$ corrupted by additive noise:
\(
    x_i = x_i^* + \eta_i, \quad x_i^+ = x_i^{+*} + \eta_i^+,
\)
where $\eta_i, \eta_i^+ \sim \mathcal{N}(0, \sigma^2 I)$ are independent Gaussian noise vectors.
\end{assumption}

\begin{assumption}[Lipschitz Continuity]
The representation function $f_\theta(\cdot)$ is Lipschitz continuous with constant $L > 0$, such that:
\(
    \|f_\theta(x) - f_\theta(y)\| \leq L \|x - y\|, \quad \forall x, y.
\)
\end{assumption}

As demonstrated in Proposition~\ref{prop:contrastive_learning}, contrastive learning can optimize representations under noisy data~\cite{contrastive_2021} by bounding the expected representation distance for clean positive pairs \( x_i^* \) and \( x_i^{+*} \).
By reducing the expectation of noisy pair distance to a constant, it ensures that the clean data representation distance is also optimized:

\begin{proposition}
\label{prop:contrastive_learning}
Contrastive learning can reduce the impact of noise on the embedding space. Let \( f_\theta(\cdot) \) denote the representation function parameterized by \(\theta\), and let \( x_i \) and \( x_i^+ \) be noisy samples derived from clean signals \( x_i^* \) and \( x_i^{+*} \). 
For Gaussian white noise, the expected bound on the representation distance for clean pairs is:
\[
\mathbb{E}_{\eta}[\|f_\theta(x_i^*) - f_\theta(x_i^{+*})\|] \leq C + 2L \sqrt{d} \sigma,
\]
where \(\mathbb{E}_{\eta}\) represents the expectation over the Gaussian noise distribution, \( C \) is the minimal representation distance for clean pairs, \( L \) is the Lipschitz constant of \( f_\theta \), \( d \) is the dimensionality of the input data, and \( \sigma \) is the standard deviation of the Gaussian noise.
\end{proposition}

To achieve these objectives, we first preprocess the input time series data for contrastive learning and reconstruction (as shown in Figure~\ref{fig:arch} (A)).
Given a time series \( \mathbf{Y} \in \mathbb{R}^{C \times T} \), where \( C \) is the number of channels and \( T \) is the number of time points, we divide it into several patches, each of size \( \mathbf{p}_i \in \mathbb{R}^{C \times t_p} \), where \( t_p \) denotes the time points per patch. For each patch \( \mathbf{p}_i \), we generate a positive sample \( \mathbf{p}_i^{+} = \mathbf{p}_i + \epsilon \), where \( \epsilon \sim \mathcal{N}(0, \sigma^2) \) adds small Gaussian noise. Multiple negative samples \( \mathbf{p}_i^{-1}, \mathbf{p}_i^{-2}, \dots, \mathbf{p}_i^{-K} \) are generated by flipping \( \mathbf{p}_i \) along the time axis as \( \mathbf{p}_i^{-k} = \text{Flip}(\mathbf{p}_i) \) for \( k = 1, 2, \dots, K \). The patches are passed through the embedding layer \( F(x) \), producing embeddings \( \mathbf{X}_i \in \mathbb{R}^{D} \), where \( D \) is the embedding dimension:
\(
\mathbf{X}_i = F(\mathbf{p}_i), \quad \mathbf{X}_i^{+} = F(\mathbf{p}_i^{+}),
\quad \mathbf{X}_i^{-k} = F(\mathbf{p}_i^{-k}), \quad k = 1, 2, \dots, K.
\) To optimize these embeddings, we employ the InfoNCE loss, which encourages positive pairs to have similar embeddings and separates negative pairs~\cite{infoNCE_2018}. The InfoNCE loss encourages the model to minimize the distance between embeddings of semantically similar (noisy) inputs while pushing apart dissimilar (transformed) inputs, thus promoting noise-robust and discriminative representations. This aligns with the theoretical insight in Proposition~\ref{prop:contrastive_learning}, which shows that minimizing noisy representation distances indirectly bounds the clean representation distances.
{\small
\begin{align*}
\mathcal{L}_{\text{InfoNCE}}\!=\!-\log \frac{\exp(\text{sim}(\mathbf{X}_i, \mathbf{X}_i^{+}) / \tau)}{\exp(\text{sim}(\mathbf{X}_i, \mathbf{X}_i^{+}) / \tau)\!+\!\sum_{k=1}^{K} \exp(\text{sim}(\mathbf{X}_i, \mathbf{X}_i^{-k}) / \tau)},
\end{align*}
}

where \( \text{sim}(\cdot, \cdot) \) denotes cosine similarity, and \( \tau \) is the temperature parameter. The inclusion of \( K \) negative samples enhances the ability to learn discriminative embeddings.

To ensure invertibility, we introduce a reconstruction head \( G: \mathbb{R}^D \rightarrow \mathbb{R}^{C \times t_p} \), mapping embedding \( \mathbf{X}_i \) back to the original patch space. 
The reconstruction is denoted by \( \hat{\mathbf{p}}_i = G(\mathbf{X}_i) \), minimizing the Mean Squared Error (MSE) loss:
\(
\mathcal{L}_{\text{MSE}} = \frac{1}{N} \sum_{i=1}^N \|\mathbf{p}_i - \hat{\mathbf{p}}_i\|_2^2.
\)

\subsection{TimeFormer}
\begin{figure*}
    \centering
    \includegraphics[width=0.86\linewidth]{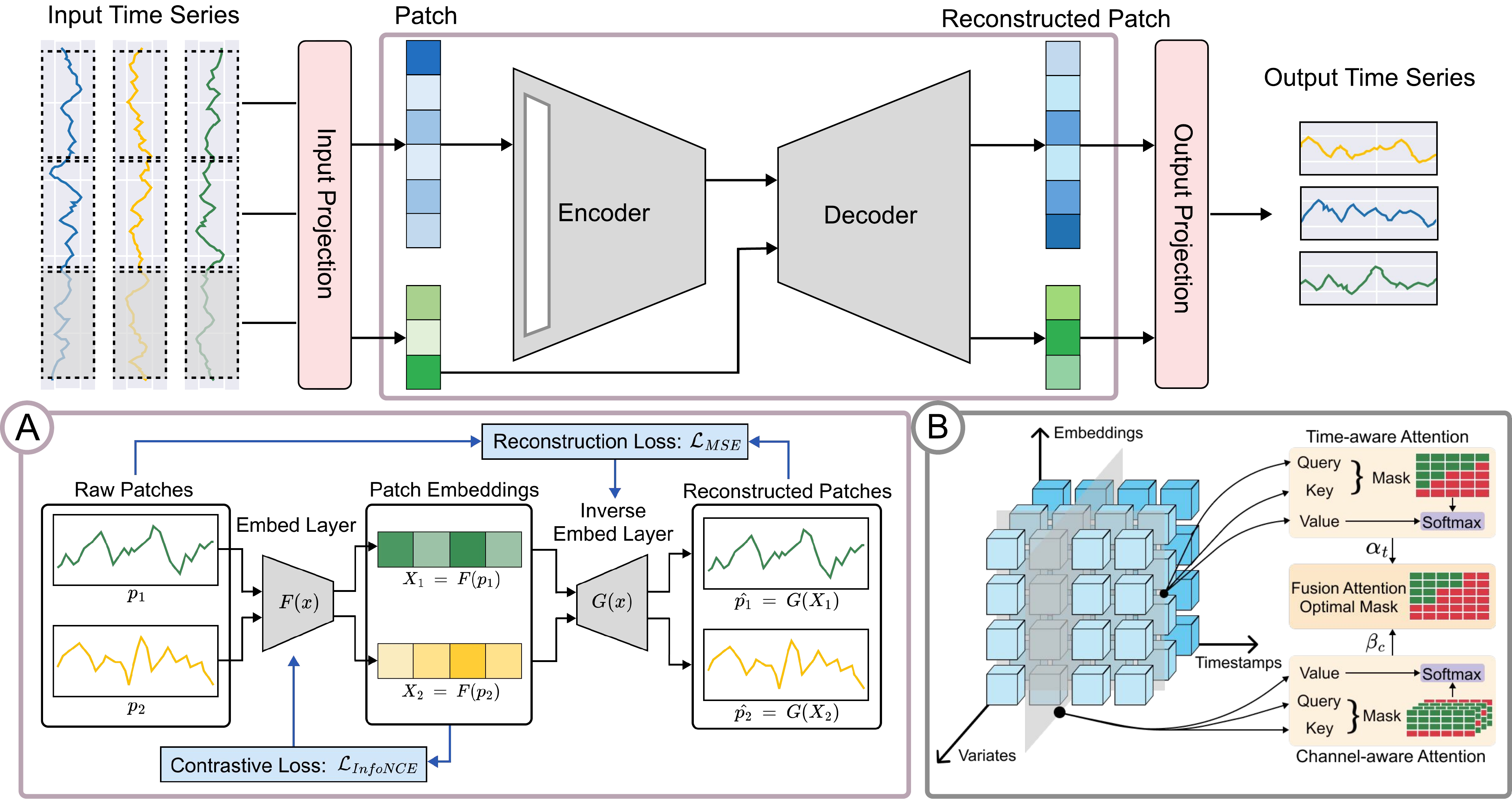}
    \vspace{-0.5em}
    \caption{The overall architecture of \textbf{LENS} is illustrated in this figure. Taking the forecasting task as an example, a 3-variate time series is visualized. The shaded patches represent the forecast horizon, whose corresponding embedding is fed into \textbf{LENs}, an encoder-decoder structure. (A) represents the invertible contrastive learning module, while (B) illustrates the attention mechanism within the TimeFormer model, comprising Time-aware attention and Channel-aware attention.}
    \vspace{-1em}
    \label{fig:arch}
\end{figure*}

\textbf{LENS} adopts an encoder-decoder architecture composed of multiple TimeFormer blocks. Each block integrates two specialized attention mechanisms—time-aware and channel-aware attention—which are designed to model temporal and inter-channel dependencies in sequential data (see Figure~\ref{fig:arch} (B)).

The attention score is computed using projected queries and keys, without flattening the sequence, thus preserving temporal structure explicitly. The main distinction between time-aware and channel-aware attention lies in their masking mechanisms. For time-aware attention, the attention score is computed as:\(
S_t = \frac{W_t^Q X (W_t^K X)^\top}{\sqrt{H}},
\) 
In this formula, \( W_t^Q \in \mathbb{R}^{D \times H} \) and \( W_t^K \in \mathbb{R}^{D \times H} \) are trainable weight matrices, \( X \in \mathbb{R}^{B \times T \times D} \) is the input feature matrix (with \( B \) being the batch size, \( T \) the temporal dimension, and \( D \) the embedding dimension), and \( H \) is the projection dimension. The resulting attention score \( S_t \) has dimensions \( \mathbb{R}^{B \times T \times T} \).

For both time-aware and channel-aware attention, mask matrices \( M_t \) and \( M_c \) are defined as follows:
\[
M_t(c, t, c', t') = 
\begin{cases} 
1, & \text{if } c = c' \text{ and } t' \leq t, \\
0, & \text{otherwise.}
\end{cases}
\]
\[
M_c(c, t, c', t') = 
\begin{cases} 
1, & \text{if } t = t' \text{ and } c \neq c', \\
0, & \text{otherwise.}
\end{cases}
\]
Here, \( c \) and \( c' \) denote channel indices, while \( t \) and \( t' \) denote time indices.
The time-aware attention mask computes attention only within the same channel, respecting temporal causality (i.e., future time steps cannot attend to past). 
Meanwhile, the channel-aware attention mask allows interactions solely between different channels at the same timestep.
With masks applied, the final time-aware attention output is computed as:\(
A_t \leftarrow \textit{softmax}\left(S_t \odot M_t\right) V_t,
\)
where \( M_t \in \mathbb{R}^{T \times T} \) is the mask matrix, and \( V_t \in \mathbb{R}^{B \times T \times D} \) is the value matrix. Finally, the outputs of time-aware and channel-aware attention are fused through a weighted sum, with \( \alpha_{\text{t}} \) and \( \beta_{\text{c}} \) being learnable scalar parameters: \(
\textit{A}_{\text{f}} = \alpha_{\text{t}} \cdot \textit{A}_{\text{t}} + \beta_{\text{c}} \cdot \textit{A}_{\text{c}}.
\)

Theoretical analysis shows that time-aware and channel-aware attention mechanisms reduce the overall model complexity by independently modeling the time and channel dimensions. 
This separation plays a key role in enhancing the model's ability to generalize, particularly in noisy environments, by mitigating the impact of noise on the learning process, as demonstrated in Proposition~\ref{prop:generalization_error_bounds}.

To formalize this, we extend the generalization error bounds for traditional and fusion attention by considering noisy input data.

\begin{definition}[Noisy Data]
Let the input data \(\mathbf{X}\) be composed of a clean signal \(\mathbf{X}_{\text{true}}\) and an additive noise component \(\mathbf{E}\):
\(
\mathbf{X} = \mathbf{X}_{\text{true}} + \mathbf{E},   
\)
where \(\mathbb{E}[\mathbf{E}] = 0\) and \(\mathbb{E}[\|\mathbf{E}\|^2] = \sigma^2\), with \(\sigma^2\) denoting the noise variance.
\end{definition}

\begin{assumption}[Bounded Loss]
The loss function \(\mathcal{L}(f(\mathbf{X}), y)\) is assumed to be bounded by a constant \(L_{\mathcal{L}}\):
\(
   \mathcal{L}(f(\mathbf{X}), y) \leq L_{\mathcal{L}}.
\)
\end{assumption}

\begin{theorem}[Generalization Error Bound via Rademacher Complexity with Noise]
For a hypothesis class \(\mathcal{F}\) and noisy input \(\mathbf{X} = \mathbf{X}_{\text{true}} + \mathbf{E}\), the generalization error \(\mathcal{E}_{\text{gen}}\) is bounded as follows:
\begin{equation*}
\mathcal{E}_{\text{gen}} \leq 2 \hat{\mathcal{R}}_n(\mathcal{F}) + \sigma L_{\mathcal{L}} + \sqrt{\frac{\log(1/\delta)}{2n}},
\end{equation*}
where \(\hat{\mathcal{R}}_n(\mathcal{F})\) is the empirical Rademacher complexity of the hypothesis class, \(\sigma\) is the noise standard deviation, \(L_{\mathcal{L}}\) bounds the impact of noise on the loss, \(n\) is the sample size, and \(\delta\) is the confidence level.
\end{theorem}

\begin{lemma}[Rademacher Complexity of Attention Mechanisms with Noisy Data]
Let \(\mathcal{F}_{\text{global}}\) and \(\mathcal{F}_{\text{fusion}}\) represent the hypothesis spaces of the traditional attention and fusion attention mechanisms, respectively. 
The Rademacher complexities of these mechanisms, considering the noise, are bounded as:
\begin{equation*}
\hat{\mathcal{R}}_n(\mathcal{F}_{\text{global}}) \leq \sqrt{\frac{\|\Sigma_{(T \cdot C)}\|^2}{n}} + \sigma \|\Sigma_{(T \cdot C)}\|,
\end{equation*}
\begin{equation*}
\hat{\mathcal{R}}_n(\mathcal{F}_{\text{fusion}}) \leq \sqrt{\frac{\alpha_t^2 \|\Sigma_T\|^2 + \beta_c^2 \|\Sigma_C\|^2}{n}} + \sigma \sqrt{\alpha_t^2 \|\Sigma_T\|^2 + \beta_c^2 \|\Sigma_C\|^2},
\end{equation*}
where $\Sigma$ terms are covariance metrics, \(\alpha_t\) and \(\beta_c\) are learnable parameters controlling the weights of each attention head.
\end{lemma}

\begin{proposition}[Comparison of Generalization Error Bounds with Noise]
\label{prop:generalization_error_bounds}
Given the covariance matrix relationships \( \|\Sigma_T\| \ll \|\Sigma_{(T \cdot C)}\| \) and \( \|\Sigma_C\| \ll \|\Sigma_{(T \cdot C)}\| \), the Rademacher complexity of the fusion attention mechanism satisfies:
\[
\hat{\mathcal{R}}_n(\mathcal{F}_{\text{fusion}}) \leq \hat{\mathcal{R}}_n(\mathcal{F}_{\text{global}}).
\]
Consequently, the generalization error of the fusion attention mechanism is bounded as:
{\small
\begin{align*}
\mathcal{E}_{\text{gen}}^{\text{fusion}} \leq \mathcal{E}_{\text{gen}}^{\text{global}} + \sigma \left( \sqrt{\alpha_t^2 \|\Sigma_T\|^2 + \beta_c^2 \|\Sigma_C\|^2}\!-\!\|\Sigma_{(T \cdot C)}\| \right),
\end{align*}}
where the second term reflects the impact of noise on the generalization performance. 
This result indicates that fusion attention can better generalize in noisy environments by reducing the overall complexity through independent modeling of time and channel dimensions.
\end{proposition}

The remaining part of the TimeFormer block shares the same structure as a standard Transformer block, including a feedforward layer and residual connections.

\subsection{Training Process}
The training process for the \textbf{LENS} begins with the invertible embedding module, which maps raw time series patches into a latent space and then reconstructs them back into the original space. 
The process involves generating positive and negative samples per patch, which are then embedded using the embedding layer. 
The model is trained using two key loss functions: the contrastive loss encourages the model to bring similar patches closer in the latent space while pushing dissimilar ones apart. The reconstruction loss ensures that the patches can be accurately reconstructed from their latent representations. Once the embedding module achieves satisfactory performance, it is frozen, meaning its weights are no longer updated during subsequent training stages.

In the second phase, the model undergoes multi-task pretraining focused on the TimeFormer Encoder and Decoder.
This phase involves training the model on a range of tasks varying input-output time series lengths (e.g., 96 $\rightarrow$ 720, 96 $\rightarrow$ 360, 48 $\rightarrow$ 360, 48 $\rightarrow$ 180).
Each task splits the input time series into context and prediction patches, embeds them via the frozen embedding module.
The TimeFormer Encoder processes the encoded context, and the Decoder generates predictions using both the encoded context and the prediction patches in a process known as teacher forcing. This approach prevents the accumulation of errors in the autoregressive decoding process. The model is trained by minimizing the MSE loss between the predicted and actual time series patches, ensuring accurate future predictions across different tasks.

\section{Experiments}
\subsection{Financial Data for Pretraining}
We collect a large-scale financial time series dataset for pre-training \textbf{LENS}, encompassing 19 sub-datasets that capture a broad spectrum of financial data.
Each sub-dataset may consist of multiple related time series, offering a comprehensive view.
This dataset covers a wide range of sampling frequencies, from macro intervals like yearly data to more granular intervals, such as seconds, with over 100 billion observations in total. However, curating such large-scale financial datasets presents challenges due to the lack of comprehensive curation across various types and frequencies, despite the abundance of available financial time series data.

To characterize the inherent complexity of each sub-dataset, we analyze statistical metrics, including Augmented Dickey-Fuller (ADF) test statistics~\cite{adf_1992}, forecastability~\cite{forecastable_2013}, and Hurst exponent~\cite{hurst_2004}. 
In datasets containing series of varying lengths, we implement length-weighted versions to ensure each time series contributes proportionally to the overall indicators.
The length-weighted ADF, forecastability, and Hurst exponent are calculated as follows: 
\[
\begin{aligned}
T &= \sum_{i=1}^C T_i, ~~
\text{ADF}(\mathcal{D}) = \sum_{i=1}^C \frac{T_i}{T} \text{ADF}(\mathbf{S}^{(i)}), \\
\text{Fore}(\mathcal{D}) &= \sum_{i=1}^C \frac{T_i}{T} (1 - \text{Entropy}(\mathcal{F}(\mathbf{S}^{(i)}))),
\end{aligned}
\]
where $\mathbf{S}_i \in \mathbb{R}^{T_i}$ denotes the $i$-th time series in dataset $\mathcal{D}$, $T_i$ is the length of $\mathbf{S}_i$, and $C$ is the number of time series in dataset $\mathcal{D}$. $\mathcal{F}(\mathbf{S}^{(i)})$ denotes the Fourier decomposition of series $\mathbf{S}^{(i)}$. 

The Dataset exhibits remarkable diversity, as indicated by the wide range of values across these metrics. The ADF test statistics show a broad distribution, from extremely negative to positive, suggesting various stationary characteristics within the dataset. Notably, many sub-datasets have low length-weighted forecastability values (less than 0.2), implying a higher degree of randomness, noise, or complex nonlinear relationships, presenting a challenge for models. The length-weighted Hurst exponent distribution reveals that most time series display long-term dependencies, which pose additional challenges for \textbf{LENS} in effectively capturing and modeling these dependencies. Further details on the methodology, key statistics, and analysis of this dataset can be found in Table~\ref{tab:dataset}.

\begin{table*}[th]
\caption{Detailed descriptions of the collected Financial Time Series Dataset. \textbf{Freq.} denotes the sampling interval of observations: S stands for a second interval; T for a minute interval; D for a daily interval; W for a weekly interval; M for a monthly interval; Q for a quarterly interval; and Y for an annual interval. \textbf{Time Series} represents the number of time series. \textbf{Obs.} denotes the total number of time points aggregating from all variates. \textbf{ADF.} denotes the Augmented Dickey-Fuller test statistics of the dataset. \textbf{Forecast.} denotes the length-weighted forecastability of the dataset, while \textbf{Hurst} denotes its length-weighted Hurst exponent.}
\vspace{-2mm}
\centering
\resizebox{0.85\textwidth}{!}{%
\begin{tabular}{ccccccc}
\hline
    \textbf{Dataset} & \textbf{Freq.} & \textbf{Time Series} & \textbf{Obs.} & \textbf{ADF.} & \textbf{Forecast.} & \textbf{Hurst}  \\ \hline
    TAQ quotes & S & 36720 & 17489815200 & -23.04 & 0.08 & 0.92 \\
    TAQ trades & S & 38952 & 136216080000 & -35.96 & 0.13 & 0.79 \\
    High frequency stock price - 1min/2min & T/2T & 4380+7065 & 492015036+394393734 & -4.22 & 0.60 & 0.94 \\
    SP500 index & D & 8 & 3321488 & -9.62 & 0.15 & 0.77 \\
    SP500 spindx\_sprtm & D & 2 & 241340  & -8.24 & 0.12 & 0.76 \\
    SP500 monthly index & M & 10 & 166480 & -3.30 & 0.13 & 0.82 \\
    Portfolio on SP 500 & D & 20 & 461670 & -5.77 & 0.13 & 0.78 \\
    Stock file index & D & 136 & 1858848 & -6.12 & 0.16 & 0.82 \\
    CRSP daily price 1984-2024 & D & 494883 & 118664274 & -13.07 & 0.21 & 0.59 \\ 
    Option & D & 4202424 & 7650299467 & -1.79 & 0.23 & 0.96 \\ 
    Future Series & D & 75192 & 339635380 & -2.26 & 0.55 & 0.98 \\
    Commodity close price & W & 311 & 2907539 & -2.48 & 0.30 & 0.95 \\
    Portfolio monthly results & M & 408 & 470016 &  -0.70 & 0.17 & 0.88 \\
    Financial ratio firm level & M & 8352 & 28440504 & -2.70 & 0.24 & 0.90 \\
    Financial ratio industry level & M & 7152 & 5759494 & -3.13 & 0.27 & 0.88 \\
    Portfolio quarterly rebalancing & Q & 90 & 34470 & 1.37 & 0.23 & 0.99 \\
    US treasury and inflation index & Q & 20 & 6540 & -3.61 & 0.16 & 0.88 \\
    CRSP daily portfolio statistics & Y & 466 & 27101 & -2.60 & 0.15 & 0.90 \\
    CRSP monthly portfolio statistics & Y & 114 & 11058 & 0.44 & 0.13 & 0.97 \\ \hline
\end{tabular}
}
\vspace{-0.5em}
\label{tab:dataset}
\end{table*}

\begin{table*}[h]
\caption{Multivariate long-term forecasting results. We use prediction lengths $T \in \{96, 196, 336, 720\}$ with a fixed context length 96. Various model architectures are included, including MLP-based (TimeMixer, DLinear), CNN-based (TimesNet, MICN), RNN-based (SegRNN), and Transformer-based models (Transformer, Crossformer, iTransformer, PatchTST).  To ensure a fair comparison and minimize the influence of parameter count differences, the small version of \textbf{LENS} is used. The best results for each prediction length are highlighted in bold, with the second-best results underlined. Percentage improvements achieved by \textbf{LENS} over the best-performing baseline model are shown in red.}
\vspace{-3mm}
\centering
\resizebox{0.84\textwidth}{!}{%
    \begin{tabular}{ccccccccccc}
    \toprule
    	& \multicolumn{4}{c}{\textbf{MLP}} & \multicolumn{4}{c}{\textbf{CNN}} & \multicolumn{2}{c}{\textbf{RNN}} \\
    	\cmidrule(lr){2-5} \cmidrule(lr){6-9} \cmidrule(lr){10-11} 
    	& \multicolumn{2}{c}{\textbf{TimeMixer}} & \multicolumn{2}{c}{\textbf{DLinear}} & \multicolumn{2}{c}{\textbf{TimesNet}} & \multicolumn{2}{c}{\textbf{MICN}} & \multicolumn{2}{c}{\textbf{SegRNN}} \\
    	\midrule
    	& \textbf{MSE} & \textbf{MAE} & \textbf{MSE} & \textbf{MAE} & \textbf{MSE} & \textbf{MAE} & \textbf{MSE} & \textbf{MAE} & \textbf{MSE} & \textbf{MAE} \\
    	\midrule
    96  & 1.0615 & 0.5232 & 2.1670 & 0.8125 & 1.4969 & 0.6510 & 2.387 & 0.9872 &\underline{1.0038} & 0.5301 \\
    192 & 2.2325 & 0.7906 & 4.8563 & 1.2315 & 2.4293 & 0.8446 & 6.3561 & 0.5380 & 2.0533 & 0.7768 \\
    336 & 4.1985 & 1.0952 & 10.9375 & 1.8197 & 3.7354 & 1.0683 & 12.1203 & 2.0467 & 5.4385 & 1.3483  \\
    720 & 7.2513 & \underline{1.4742} & 25.1411 & 2.8695 & \underline{7.1981} & \underline{1.4785} & 26.5799 & 3.0608 & 8.1561 & 1.6285  \\
    
    \midrule
    
           & \multicolumn{8}{c}{\textbf{Transformer}} & \multicolumn{2}{c}{\multirow{2}{*}{\textbf{LENS\textsubscript{S}}}} \\
            \cmidrule(lr){2-9} 
           & \multicolumn{2}{c}{\textbf{Transformer}} & \multicolumn{2}{c}{\textbf{Crossformer}} & \multicolumn{2}{c}{\textbf{iTransformer}} & \multicolumn{2}{c}{\textbf{PatchTST}}  \\
           \midrule
           & \textbf{MSE} & \textbf{MAE} & \textbf{MSE} & \textbf{MAE} & \textbf{MSE} & \textbf{MAE} & \textbf{MSE} & \textbf{MAE} & \textbf{MSE} & \textbf{MAE} \\
            \midrule
    
    \textbf{96} & 37.9486 & 3.8529 & 28.3023 & 3.0753 & 1.2629 & 0.5870 & 1.0079 & \underline{0.5151} & \textbf{0.4043} \textcolor{red}{(-59.72\%)}   & \textbf{0.4654} \textcolor{red}{(-9.65\%)}  \\
    \textbf{192} & 37.9185 & 3.8589 & 30.3626 & 3.2360 &\underline{1.9356} & \underline{0.7386} & 2.0353 & 0.7564 & \textbf{0.7217} \textcolor{red}{(-62.71\%)}  & \textbf{0.6236} \textcolor{red}{(-15.56\%)}  \\
    \textbf{336} & 37.9894 & 3.8808 & 33.2249 & 3.4259 & 3.7736 & 1.0575 & \underline{3.6823} & \underline{1.0340} & \textbf{1.2038} \textcolor{red}{(-67.31\%)}   & \textbf{0.8295} \textcolor{red}{(-19.78\%)}  \\
    \textbf{720} & 39.6535 & 3.9306 & 35.8180 & 3.6033 & 8.1084 & 1.5758 & 7.7411 & 1.5241& \textbf{2.8034} \textcolor{red}{(-61.05\%)} & \textbf{1.2316} \textcolor{red}{(-16.70\%)} \\
    \bottomrule
\end{tabular}
}
\vspace{-3mm}
\label{tab:forecasting}
\end{table*}

\begin{table*}[h]
\caption{Imputation results. The used masked ratios are set at \{0.125, 0.25, 0.375, 0.5\}. Selected models are representative and well-suited for this task. For \textbf{LENS}, the encoder module was decoupled for training specific to this task. Best results are highlighted in bold, and the second-best results are underlined. Percentage improvements achieved by \textbf{LENS} over the best-performing baseline model are shown in red.}
\vspace{-2mm}
\resizebox{0.84\textwidth}{!}{%
\begin{tabular}{@{}ccccccccccccc@{}}
\toprule
& \multicolumn{2}{c}{\textbf{Transformer}} & \multicolumn{4}{c}{\textbf{MLP}} & \multicolumn{2}{c}{\textbf{CNN}} & \multicolumn{2}{c}{\multirow{2}{*}{\textbf{LENS\textsubscript{S}}}} \\
        \cmidrule(lr){2-3} \cmidrule(lr){4-7} \cmidrule(lr){8-9}
        & \multicolumn{2}{c}{\textbf{iTransformer}} & \multicolumn{2}{c}{\textbf{TiDE}} & \multicolumn{2}{c}{\textbf{DLinear}} & \multicolumn{2}{c}{\textbf{MICN}} & & \\
        \midrule
       & \textbf{MSE} & \textbf{MAE} & \textbf{MSE} & \textbf{MAE} & \textbf{MSE} & \textbf{MAE} & \textbf{MSE} & \textbf{MAE} & \textbf{MSE} & \textbf{MAE} \\
       \midrule
0.125 & \underline{0.2284} &  \textbf{0.2434} & 0.3049 & 0.2822 & 0.2968 & \underline{0.2788} & 0.8399 & 0.4142 & \textbf{0.1374} \textcolor{red}{(-39.74\%)}& 0.2838 \textcolor{green}{(+16.79\%)}\\
0.250 & \underline{0.3642} & \textbf{0.3079} & 0.4896 & 0.3545 & 0.4784 & \underline{0.3523} & 2.4919 & 0.8036 &\textbf{0.2606} \textcolor{red}{(-28.41\%)}& 0.4068 \textcolor{green}{(+32.08\%)}\\
0.375 & \underline{0.5161} & \textbf{0.3662} & 0.6814 & 0.4174 & 0.6480 & \underline{0.4106} & 4.8870 & 1.1210 &  \textbf{0.4552} \textcolor{red}{(-11.78\%)} & 0.5529 \textcolor{green}{(+51.07\%)} \\
0.500 & \underline{0.7268} & \textbf{0.4339} & 0.9931 & 0.5050 & 0.8865 & \underline{0.4780} & 7.9811 & 1.5730 & \textbf{0.7001} \textcolor{red}{(-3.70\%)}&  0.7034 \textcolor{green}{(+62.07\%)}  \\
	\bottomrule
\end{tabular}	
}
\vspace{-0.2em}

\label{tab:imputation}
\end{table*}

\begin{table*}[h]
\caption{Portfolio management results. `LB-FW' means the lengths of lookback and forward. Best results are highlighted in \textbf{bold}, and the second best results are \underline{underlined}.}
\centering
\vspace{-2mm}
\resizebox{\textwidth}{!}{%
\begin{tabular}{ccccccccccccccccccc}
        \toprule	
        \multirow{2}{*}{\textbf{LB-FW}} & \multicolumn{3}{c}{\textbf{Equal Weighting}} & \multicolumn{3}{c}{\textbf{Market Cap Weighting}} &  \multicolumn{3}{c}{\textbf{Volatility Weighting}} & \multicolumn{3}{c}{\textbf{Markowitz Model}} & \multicolumn{3}{c}{\textbf{Min-Variance Weighting}} &  \multicolumn{3}{c}{\textbf{LENS\textsubscript{S}}} \\
        \cmidrule(lr){2-4} \cmidrule(lr){5-7} \cmidrule(lr){8-10} \cmidrule(lr){11-13} \cmidrule(lr){14-16} \cmidrule(lr){17-19}
        & $R_d$ & $S_a$ & MDD & $R_d$ & $S_a$ & MDD & $R_d$ & $S_a$ & MDD & $R_d$ & $S_a$ & MDD & $R_d$ & $S_a$ & MDD & $R_d$ & $S_a$ & MDD\\
        \midrule
        \textbf{100-5} & 0.0004 & 1.1794 & -0.0242 & 0.0002 & 1.5079 & -0.0113 & 0.0003 & 1.5277 & -0.0106 & 0.0001 & 1.5172 & -0.0096 & 0.0001 & 1.5207 & -0.0096 & 0.0005& 1.4658 & -0.0212 \\
        \textbf{100-10} & 0.0004 & 1.1794 & -0.0242 & 0.0002 & 1.1515 & -0.0243 & 0.0003 & 1.1676 & -0.0227 & 0.0001 & 1.1115 & -0.0208 & 0.0001 & 1.1175 & -0.0208 & 0.0005 & 1.1681 & -0.0433\\
        \textbf{255-10} & 0.0004 & 1.1902 & -0.0242 & 0.0002 & 1.1631 & -0.0244 & 0.0003 & 1.1577 & -0.0229 & 0.0001 & 1.0800 & -0.0206 & 0.0001 & 1.0811 & -0.0206 & 0.0004 & 1.1225 & -0.0446 \\
        \textbf{255-20} & 0.0004 & 1.0054 & -0.0403 & 0.0002 & 0.9597 & -0.0410 & 0.0003 & 0.9550 & -0.0382 & 0.0001 & 0.8862 & -0.0348 & 0.0001 & 0.8861 & -0.0348 & 0.0004 & 1.1091 & -0.0695\\
        \bottomrule
\end{tabular}
}
\vspace{-3mm}
\label{tab:portofolio}
\end{table*}

\subsection{Downstream Tasks and Baseline}
\begin{figure}[ht]
    \centering
    \includegraphics[width=0.88\columnwidth]{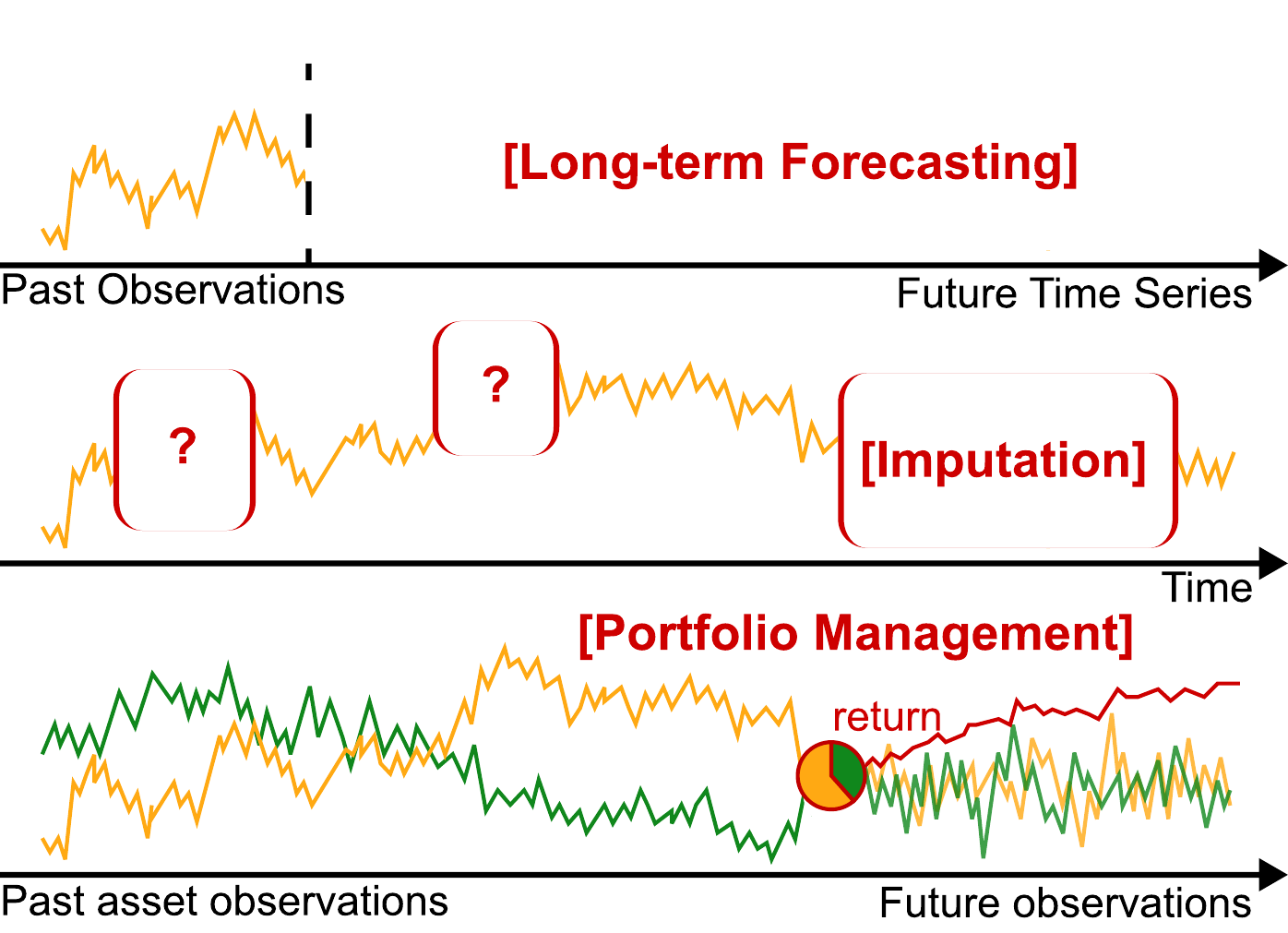}
    \vspace{-2mm}
    \caption{Schematic illustration of three financial time series analysis tasks.}
    \label{fig:task}
    \vspace{-4mm}
\end{figure}

We evaluate \textbf{LENS} on three fundamental but challenging tasks in financial time series analysis, as illustrated in Figure~\ref{fig:task}.
Long-term forecasting and imputation are also common to general time series, while portfolio management is specific to financial time series data.

\paragraph{Long-term Forecasting.}
The goal of forecasting is to predict future time points based on the input time series.
We report the Mean Square Error (MSE) and Mean Absolute Error (MAE), comparing against nine state-of-the-art baselines, Transformer~\cite{transformer_2017}, Crossformer~\cite{crossformer}, iTransformer~\cite{itransformer_2023}, PatchTST~\cite{patchtst}, TimeMixer~\cite{timemixer}, DLinear~\cite{Dlinear}, TimesNet~\cite{timesnet}, MICN~\cite{micn}, and SegRNN~\cite{segrnn}.
These baselines can be categorized into four groups based on their backbone architecture, including Transformer-based, MLP-based, CNN-based, and RNN-based.
MSE and MAE are calculated as follows:
$MSE = \frac{1}{n}\sum_{i=1}^n(y_i-\hat{y_i})^2,~~ MAE = \frac{1}{n}\sum_{i=1}^n|y_i-\hat{y_i}|,$
where $y, \hat{y} \in \mathbb{R}^{F \times C}$ represent the ground truth and predicted results, respectively, for $F$ future time points and $C$ dimensions.
$y_i$ denotes the $i$-th future time point.
The lookback sequence length is set to 96, and the prediction lengths $T$ considered include $\{96, 196, 336, 720\}$.

\vspace{-3mm}
\paragraph{Imputation.}
The goal of time-series imputation is to recover the value of missing time points precisely.
We also report MSE and MAE metrics on the imputation task, comparing against four baselines: iTransformer~\cite{itransformer_2023}, TiDE~\cite{TiDE}, Dlinear~\cite{Dlinear}, and MICN~\cite{micn}, covering Transformer-based, MLP-based, and CNN-based.
For this task, the lookback sequence length is set to $96$, and the top-k value is set to 5, meaning the top 5 most relevant candidate values are considered during the imputation process. 
The masked ratios are set at $\{0.125, 0.25, 0.375, 0.5\}$, allowing for a robust assessment.

\vspace{-3mm}
\paragraph{Portfolio Management.}
Portfolio management involves the selection and optimization of asset allocation to maximize the total (or average) return within a given investment process~\cite{RL_fpm_2019}.
In the case of portfolio management, we choose the following metrics:
(1) Simple daily return ($R_d$) measures the return of an asset over one day, calculated as: $R_d = \frac{P_t-P_{t-1}}{P_{t-1}}$, where $P_t$ is the asset price at time $t$ and $P_{t-1}$ is the asset price at the previous trading day.
(2) The simple annual sharpe ratio ($S_a$) measures the performance of an investment compared to a risk-free asset, calculated as $S_a = \frac{\bar{R}_a - R_f}{\sigma_a},$ where $\bar{R}_a$ is the average annual return of the portfolio, $R_f$ is the risk-free rate, and $\sigma_a$ is the standard deviation of the annual return.
(3) Maximum Drawdown (MDD) measures the maximum loss from a peak to a trough of an asset's price, before a new peak is attained.
It is defined as $MDD=\mathop{\max}_{t \in [1,T]} (\frac{\max_{j \in [1,t]} P_j-P_t}{\max_{j \in [1,t]} P_j}),$ where $P_t$ is the asset price at time $t$, and $T$ is the total time period considered.

We compare \textbf{LENS} with five common portfolio management strategies~\cite{cap_weighted_portfolio}: equal weighting, market cap weighting, volatility weighting, minimum variance weighting, and the Markowitz mean-variance 
model~\cite{MMVM_2000}.
Equal weighting assigns equal weights to all assets. Market cap weighting allocates weights based on market capitalization. Volatility weighting assigns higher weights to less volatile assets to reduce risk.
Minimum variance minimizes portfolio volatility by optimizing the covariance matrix of asset returns. The Markowitz model balances expected return and risk, ensuring non-negative weights that sum to one for optimal asset allocation.
In each downstream task, we compare \textbf{LENS} with state-of-the-art models, spanning different backbone architectures.
The used dataset is derived from the `CRSP daily price 1984-2024' subdataset, which includes 20 variables with 203,860 observations.
To present data leakage, this dataset is excluded from the pre-training stage.

\vspace{-1mm}
\subsection{Result Analysis}
In long-term financial time series forecasting tasks, simpler models like MICN~\cite{micn}, a CNN-based model, experience a significant increase in error as the sequence length grows. In contrast, models capable of capturing both local and global temporal patterns, such as TimeMixer (MLP-based) and TimesNet (CNN-based), exhibit more stable performance across various sequence lengths. These models offer a clear advantage in handling long-term dependencies in time series data, making them better suited for extended forecasting horizons. As demonstrated in Table~\ref{tab:forecasting}, TimeMixer outperforms DLinear among MLP-based models. TimeMixer effectively identifies distinct patterns in time series at different sampling scales and introduces an MLP-based multiscale mixing architecture~\cite{benchmark_2024}. We can also conclude that pre-training markedly enhances Transformer performance: the standard Transformer suffers from rising errors with longer sequences, while the pre-trained \textbf{LENS} maintains superior accuracy across horizons, highlighting the benefit of large-scale financial pre-training.

Imputation is another key task in time series analysis. As shown in Table~\ref{tab:imputation}, \textbf{LENS\textsubscript{S}} consistently outperforms models such as Crossformer, TiDE, and MICN. Its autoregressive pre-training allows the model to capture the intrinsic temporal structure of financial sequences, leading to more accurate and coherent reconstructions. In contrast, iTransformer produces more uniform but less precise predictions, indicating stability at the expense of structural fidelity.

Beyond basic forecasting tasks, \textbf{LENS}-based predictions also translate into tangible gains in portfolio management. As shown in Table~\ref{tab:portofolio}, portfolios constructed from its forecasts achieve higher average returns and lower risk exposure compared to both traditional models and standard statistical baselines. Although the Sharpe ratio is comparable to that of the Markowitz model, \textbf{LENS} exhibits a smaller maximum drawdown, suggesting that its accurate time-series predictions help mitigate downside risk and enhance the stability of investment strategies.

\paragraph{Ablation Study.} The ablation results in Table~\ref{tab:patch} confirm that each component of \textbf{LENS} contributes meaningfully to overall performance. The encoder and decoder prove indispensable, as removing either leads to the most severe degradation across both forecasting and imputation tasks, underscoring their roles in temporal modeling and noise reduction. The channel-, time-, and fusion-attention modules further enhance representation learning by capturing complementary dimensions of financial dynamics. Finally, the invertible embedding notably improves input–latent alignment, reducing information loss and providing a stronger foundation for downstream modeling. 

\vspace{-2mm}

\begin{table}[htbp]
\caption{The ablation study results for the \textbf{LENS\(_s\)}. The tasks include prediction (forecasting the next 96 time steps based on the previous 96 steps) and imputation (masking 12.5\% data) tasks. 
Decoder-only variant uses the information preceding the masked region for imputation.
Encoder-only variant adds a linear projection at the end for 96-step output.
These setups are designed to evaluate the impact of each component on the overall model performance in these specific tasks and to ensure a relatively fair ablation comparison.
}
\vspace{-3mm}
\centering
\resizebox{0.42\textwidth}{!}{%
\begin{tabular}{@{}cccc@{}}
\toprule
  &  & \textbf{Prediction}& \textbf{Imputation}\\
\midrule
  &{\textbf{LENS\textsubscript{S}}}  & 0.4043 & 0.1374 \\
\midrule
  & w/o Decoder & 0.8529 & 0.1755 \\
  & w/o Encoder & 1.7341 & 0.2608 \\
\midrule
  & w/o Channel Aware Attention  & 0.6073 & 0.1651 \\
  & w/o Time Aware Attention & 0.5941 & 0.1815 \\
  & w/o Fusion Attention & 0.6162 & 0.1573 \\
\midrule 
& w/o Invertible Embedding & 0.5521 & 0.1806 \\
\bottomrule
\end{tabular}	
}
\vspace{-2mm}
\label{tab:patch}
\end{table}

\vspace{-3mm}
\paragraph{Exploration of Embedding Space.} To gain deeper insights into the invertible embedding module, we utilize two metrics: alignment and uniformity~\cite{align_uni_2022,contra_metric_2022}. 
The alignment metric \( \mathcal{M}_{\text{align}} \) quantifies the proximity of the embeddings of positive pairs (an original patch \( p_i \) and its slightly perturbed version \( p_i^+ \)) in the embedding space, and is defined as:
\[
\mathcal{M}_{\text{align}}(f; \alpha) = \mathbb{E}_{(X_i, X_i^+)}\left[ \| X_i - X_i^+ \|^\alpha \right]
\]
Here, \( X_i = F(p_i) \) and \( X_i^+ = F(p_i^+) \) represent the embeddings of the original and positive patches, respectively, with \( \alpha \) set to one. The uniformity metric \( \mathcal{M}_{\text{uniform}} \) encourages the embeddings to be uniformly distributed across the hypersphere, thereby preventing them from clustering in restricted regions. It is computed as:
\[
\mathcal{M}_{\text{uniform}}(f; t) = \log \mathbb{E}_{\text{i.i.d.}} (X_i, X_j) \left[ e^{-t \| X_i - X_j \|^2} \right]
\]
where \( X_j = F(p_j) \) is another embedding from the patch set, and \( t \) is a temperature parameter. In our case, we set \( t = 1 \) as a constant.

\begin{table}[ht]
\caption{Uniformity and Alignment scores for different patch sizes (\(patch=8\) and \(patch=16\)) and embedding dimensions (\(1024\), \(2048\), \(3074\)).}
\vspace{-3mm}
\centering
\resizebox{0.45\textwidth}{!}{%
\begin{tabular}{lcccccc}
\toprule
 & \multicolumn{2}{c}{\textbf{Patch=8}} & \multicolumn{2}{c}{\textbf{Patch=16}} \\
\cmidrule(r){2-3} \cmidrule(r){4-5}
 & Uniformity & Alignment & Uniformity & Alignment \\
\midrule
\textbf{1024} & -0.1009 & 0.0066 & -0.1870 & 0.0085\\
\textbf{2048} & -0.0864 & 0.0067 & -0.1749 & 0.0084 \\
\textbf{3072} & -0.1942 & 0.0082 & -0.1745 & 0.0084 \\
\bottomrule
\end{tabular}
}
\vspace{-1mm}
\label{tab:uanda}
\end{table}

Table~\ref{tab:uanda} demonstrates that the choice of patch size and embedding dimension needs to be carefully aligned. When the patch size is 16, the alignment scores tend to decrease as the embedding dimension increases, but a bottleneck is observed. Conversely, for patch size 8, the alignment scores increase with larger dimensions, indicating that smaller patches make it easier to bring samples closer together when using contrastive loss. Additionally, the results also suggest that larger embedding dimensions may facilitate a more uniform distribution of samples across the embedding space, as reflected in the uniformity scores. Combining the insights from Table~\ref{tab:patch} and Table~\ref{tab:uanda}, it is evident that the embedding encoder-decoder structure is particularly well-suited for financial time series data. The complementary nature of the encoder and decoder allows for better representation of high-noise sequence data, enabling more effective learning and generalization. Together, these components help to extract meaningful patterns from complex and noisy financial data.

\vspace{-2mm}
\paragraph{Scaling Experiments.} We provide \textbf{LENS} in three sizes - small, base, and large, with key parameter details listed in Table~\ref{tab:size}. Models are trained on 8$\times$NVIDIA H100 (80G). 

\begin{wrapfigure}[5]{l}{0.42\columnwidth}
  \vspace{-3mm}
  \centering
  \scalebox{0.8}{
        \begin{tabular}{lcc}
            \toprule
             & \textbf{Prediction} & \textbf{Imputation} \\ 
            \midrule
            {\textbf{LENS\textsubscript{S}}} & 0.4033 & 0.1347 \\ 
            {\textbf{LENS\textsubscript{B}}} & 0.3121 & 0.1128 \\ 
            {\textbf{LENS\textsubscript{L}}} & 0.2845 & 0.1023 \\ 
            \bottomrule
        \end{tabular}
        }
        \vspace{-3mm}
        \label{tab:scale}
\end{wrapfigure}
Results show clear improvements in both prediction and imputation tasks as model size increases, as shown on the left. Specifically, the prediction task requires the model to forecast the next 96 time steps based on the previous 96 steps, while the imputation task masks 12.5\% of the data. 
This observed trend follows the fundamental scaling law, which posits that larger models, with more parameters, exhibit significantly enhanced performance. The scaling behavior underscores the importance of model size in capturing complex financial time series patterns and achieving superior accuracy.

\vspace{-3mm}
\begin{table}[h]
\caption{Configurations of \textbf{LENS} models with different model size, detailing their dimensions, number of layers, model depth, attention heads, and total parameters.}
\vspace{-3mm}
\centering
\begin{tabular}{@{}cccccc@{}}
\toprule
	& \textbf{Dims} & \textbf{Layers} & \textbf{d\textsubscript{model}} & \textbf{Heads} & \textbf{Params}\\
	\midrule
\textbf{LENS\textsubscript{S}} & 1024 & 8 & 1024 & 8 & 0.2B \\
\textbf{LENS\textsubscript{B}} & 2048 & 8 & 2048 & 16 & 1B \\
\textbf{LENS\textsubscript{L}} & 3072 & 16 & 3072 & 32 & 5B \\
	\bottomrule
\end{tabular}	
\vspace{-3mm}
\label{tab:size}
\end{table}


\vspace{-2mm}
\paragraph{Case Study.}

\begin{figure}[h!]
    \centering
    \includegraphics[width=0.86\columnwidth]{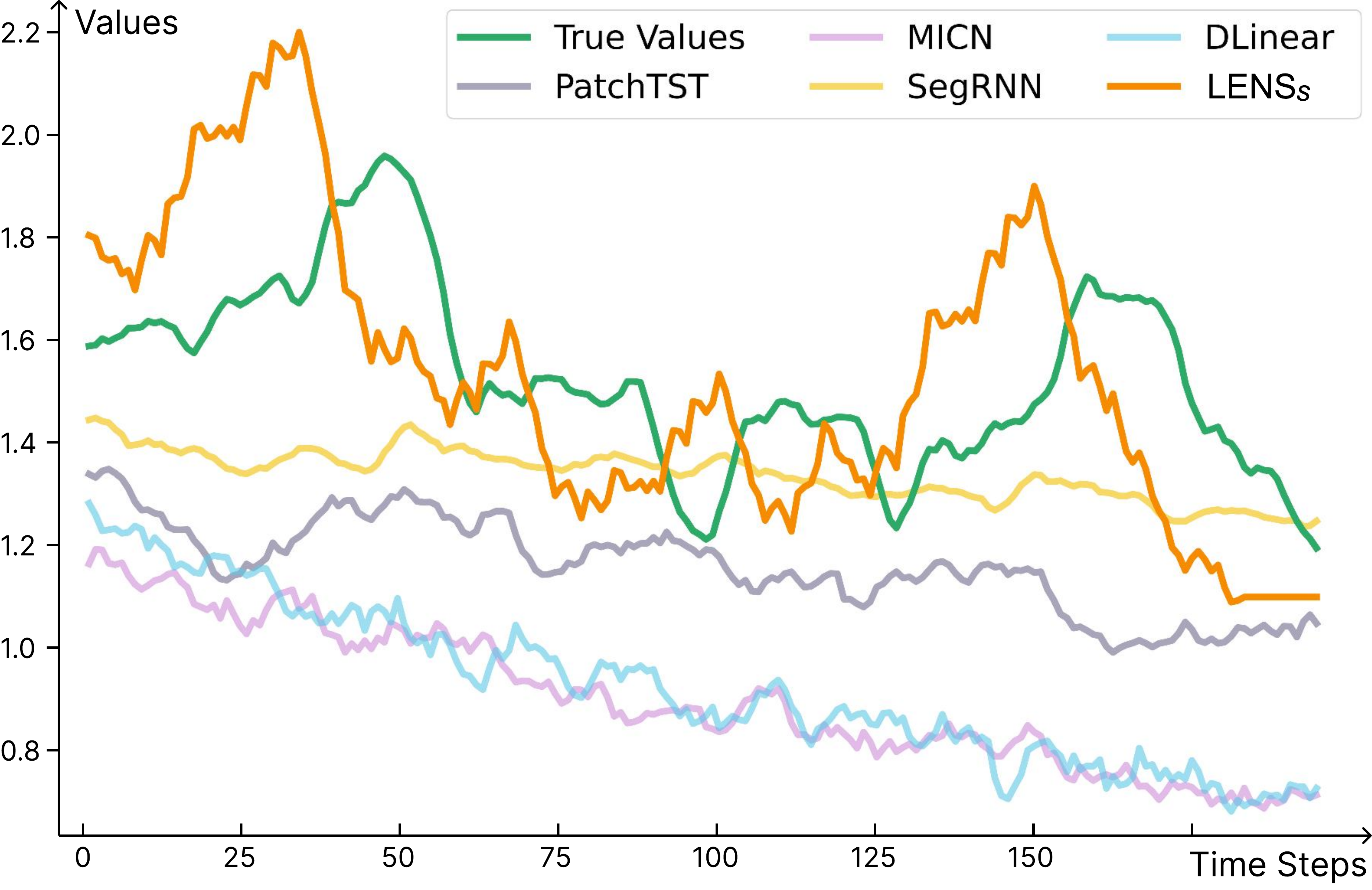}
    \vspace{-2mm}
    \caption{Forecasting Performance Comparison. This figure illustrates the forecasting results of a sample where the task involves predicting the next 196 time steps based on the previous 96 steps. The green line represents the true values, providing a reference for evaluating model performance.}
    \vspace{-3mm}
    \label{fig:sample}
\end{figure}

Figure~\ref{fig:sample} demonstrates the significant advantages of \textbf{LENS\textsubscript{S}} in predicting the next 196 time steps. Compared to other models (MICN, DLinear, PatchTST, SegRNN), \textbf{LENS\textsubscript{S}} tracks the true value curve more closely and anticipates trends effectively. 
\textbf{LENS\textsubscript{S}} can identify and predict upward or downward trends in the data ahead of time, making it more adaptive and accurate in financial time series forecasting tasks. 
Other models approximate general downward trends but struggle to accurately recognize short-term fluctuations. 
This capability of \textbf{LENS\textsubscript{S}} stems from its architecture, which effectively captures complex patterns and handles high noise, resulting in reliable predictions in practical applications.

\section{Conclusion}
In this paper, we introduce \textbf{LENS}, a pre-trained foundational model specifically tailored for financial time series analysis.
As a highly performant model, \textbf{LENS} excels at learning and capturing general patterns from large volumes of financial data, demonstrating outstanding performance in many common market scenarios and proving its substantial potential in the financial domain.
It is uniquely designed to handle time series with extremely low signal-to-noise ratios—a hallmark of financial data—enabled by an invertible embedding module and a novel attention mechanism incorporating time-aware and channel-aware attention.
Through comprehensive experiments, we validate \textbf{LENS}'s effective generalization across diverse financial downstream tasks and the contributions of its individual components.
However, \textbf{LENS} faces limitations in handling sudden market events, such as rapid increases or declines, likely due to insufficient information in the input data rather than inherent model flaws.
As the first open-source, large-scale pre-trained model for financial time series, \textbf{LENS} represents a significant step forward in addressing the challenges of financial modeling and sets the stage for future innovations, including enhancements to better incorporate abrupt market dynamics in this critical domain.
\vspace{-2mm}

\begin{acks}
Guang Zhang acknowledges the support from the Guangzhou-HKUST(GZ) Joint Funding Program (No.2024A03J0630).
\end{acks}


\bibliographystyle{ACM-Reference-Format}
\bibliography{sample-base}

\end{document}